\documentclass[11pt]{article}

\usepackage[preprint]{acl}

\usepackage{amsmath,amsfonts,bm}

\def\eqref#1{equation~\ref{#1}}

\def\1{\bm{1}}

\DeclareMathAlphabet{\mathsfit}{\encodingdefault}{\sfdefault}{m}{sl}
\SetMathAlphabet{\mathsfit}{bold}{\encodingdefault}{\sfdefault}{bx}{n}

\def\gA{{\mathcal{A}}}

\def\gM{{\mathcal{M}}}

\def\gO{{\mathcal{O}}}

\def\gS{{\mathcal{S}}}

\usepackage{times}
\usepackage{latexsym}

\usepackage[T1]{fontenc}

\usepackage{graphicx}
\usepackage{accents}
\usepackage[labelfont=bf, justification=justified,font=small]{caption}
\usepackage{wrapfig}
\usepackage{multirow}
\usepackage{xspace}
\usepackage{tikz}
\usepackage{backref}
\usepackage{enumitem} %
\usepackage{subcaption}
\usepackage{booktabs}
\usepackage{pifont}
\usepackage{listings}

\usepackage{marginnote}
\usepackage{xcolor} %

\newcommand{\myparagraph}[1]{
\vspace{0.1cm}\noindent
\textbf{#1}
}

\usepackage[utf8]{inputenc}

\usepackage{microtype}

\usepackage{inconsolata}

\usepackage{graphicx}

\title{LUMINA: Long-horizon Understanding for Multi-turn Interactive Agents}

\author{
 \textbf{Amin Rakhsha}\thanks{Work completed during an internship at Qualcomm AI Research}\textsuperscript{1} \hspace{0.15em}
 \textbf{Thomas Hehn}\textsuperscript{2} \hspace{0.15em}
 \textbf{Pietro Mazzaglia}\textsuperscript{2} \hspace{0.15em} \\
 \textbf{Fabio Valerio Massoli}\textsuperscript{2} \hspace{0.15em}
 \textbf{Arash Behboodi}\textsuperscript{2} \hspace{0.15em}
 \textbf{Tribhuvanesh Orekondy}\textsuperscript{2} \hspace{0.15em}
\\\\
 \textsuperscript{1}University of Toronto\quad
 \textsuperscript{2}Qualcomm AI Research\thanks{Qualcomm AI Research is an initiative of Qualcomm Technologies, Inc.}
}

\begin{document}
\maketitle
\begin{abstract}
Large language models can perform well on many isolated tasks, yet they continue to struggle on multi‑turn, long‑horizon agentic problems that require skills such as planning, state tracking, and long context processing.
In this work, we aim to better understand the relative importance of advancing these underlying capabilities for success on such tasks.
We develop an oracle counterfactual framework for multi-turn problems that asks: how would an agent perform if it could leverage an oracle to perfectly perform a specific task? The change in the agent's performance due to this oracle assistance allows us to measure the criticality of such oracle skill in the future advancement of AI agents.  
We introduce a suite of procedurally generated, game‑like tasks with tunable complexity.
These controlled environments allow us to provide precise oracle interventions, such as perfect planning or flawless state tracking, and make it possible to isolate the contribution of each oracle without confounding effects present in real‑world benchmarks.
Our results show that while some interventions (e.g., planning) consistently improve performance across settings, the usefulness of other skills is dependent on the properties of the environment 
and language model.
Our work sheds light on the challenges of multi‑turn agentic environments to guide the future efforts in the development of AI agents and language models.
\end{abstract}

\section{Introduction}

Large Language Models (LLMs) have demonstrated exceptional performance across a wide range of tasks, including
natural conversations \citep{touvron2023llama,achiam2023gpt}, question answering \citep{yang2018hotpotqa}, competitive coding \citep{austin2021program,white2024livebench}, and mathematical reasoning \citep{comanici2025gemini,guo2025deepseek}.
Consequently, given their general-purpose capabilities, a natural question that is actively being explored is whether language models can be leveraged as \textit{multi-turn} agents, i.e., whether they can iteratively perceive, reason, and take strategic actions towards achieving a distant goal.
Such tasks introduce a host of new challenges, such as maintaining coherence over multiple turns and long contexts, reasoning over multiple dynamic pathways, recovering from errors, identifying the right tools for the task, and efficiently tracking state and progress without explicit feedback.

Numerous benchmarks quantitatively analyze multi-turn agent capabilities across various domains, ranging from function calling \citep{patil2025berkeley}, web navigation \citep{koh2024visualwebarena}, interactive coding \citep{trivedi2024appworld}, human interaction \citep{liu2023agentbench}, and game-playing \citep{guertler2025textarena}.
Ongoing results suggest that there is plenty of progress to be made, e.g., on AppWorld \citep{trivedi2024appworld}, GPT-4o has a success rate of 30.2\% and open-weight models such as LLama3-70B \citep{grattafiori2024llama} achieve 7.0\%.
As many of the benchmark analyses report, pushing progress requires enabling numerous skills, such as efficient task decomposition, planning, state tracking, and information gathering.
Even though the necessity of these skills has been individually established in prior work, their relative importance on the agent’s success remains an open question.

In this paper, our goal is to understand which of these skills (or combination thereof) are the bottleneck to make progress towards capable multi-turn agents.
To assess skill importance, we introduce an oracle intervention framework that poses counterfactual questions:\textit{What if the agent had access to the information that skill provides?}
Using this framework, we investigate three oracle interventions and their combinations: 
\textit{planning}, which provides a description of a single-turn subtask on the optimal trajectory,
\textit{state tracking}, which provides an accurate state of the agent, and 
\textit{history pruning}, which rewrites the context to remove distractors and provide a compact summary.

Measuring the impact of individual skills on an agent’s success in multi-turn, goal-oriented tasks is challenging.
Real-world ablations are cumbersome and often impractical because these tasks may admit numerous equally optimal paths to the goal, making exhaustive oracle annotations intractable.
To overcome this, we introduce procedurally-generated game environments where optimal actions and strategies can be computed at any step, enabling systematic oracle interventions.
We consider three game-playing multi-turn tasks: 
\textit{ListWorld} to cautiously mutate a python list using only \texttt{pop} actions,
\textit{TreeWorld} to find a particular node in a tree iteratively using only \texttt{get\_children} actions, and
\textit{GridWorld} to spatially navigate a 2d grid to reach a target location while avoiding holes.
All environments are configurable, and importantly, enable us to inject accurate oracle information at any point of the agent's trajectory.

Our framework enables us to examine the capabilities of multi-turn agents along multiple dimensions (e.g., task complexity, model size, influence of specific oracle interventions) and helps us providing insights into \textbf{L}ong-horizon \textbf{U}nderstanding of \textbf{M}ulti-turn \textbf{IN}teractive \textbf{A}gents (\textbf{LUMINA}).
First, we observe a significant discrepancy with the low success rates over long-horizon trajectories in spite of high accuracy per step (i.e., whether the chosen action is from the set of optimal actions), indicating that compounding errors represent a major challenge \citep{sinha2025illusion,li2025beyond}.
Second, we find that oracle interventions generally improve the success rate, but their effectiveness varies depending on model size.
Finally, different environments have distinct characteristics that translate to varying gains from the interventions and measurable performance discrepancies between success rate and step accuracy.
Overall, our findings present a double-edged picture: while improving specific skills (enabled by oracles interventions in our case) generally help the LLM-based agents solving multi-turn tasks, fully bridging the gap likely requires exploiting environment and model-specific understanding.

\section{Related Works}

\myparagraph{LLMs and Agents.}
Agent-based systems, defined by a policy interacting with an environment to achieve goals and receive rewards \citep{russell1995modern}, have a long history.
Recent work shows that language models can serve as the policy \citep{huang2022language,yao2023react}, the environment as world model \citep{hao2023reasoning}, or the reward function \citep{zheng2023judging,zhang2024generative}.
We focus on LLMs as policies that auto-regressively sample next actions.
A prominent example is ReAct prompting \citep{yao2023react}, which interleaves reasoning and task-specific actions.
ReAct has proven effective across domains, from game playing \citep{wang2023voyager} to interactive coding \citep{trivedi2024appworld}.
In this work, we adopt ReAct-based prompting to elicit dynamic reasoning and planning from LLMs.

\myparagraph{LLM Agent Capabilities.}
The are several essential skills for agents beyond language understanding and reasoning.
Agents must execute admissible actions (e.g., tools, function calls) with correct arguments \citep{qin2023toolllm,patil2024gorilla}.
As there may be multiple paths towards the goal, agents should support multi-path reasoning \citep{besta2024graph,yao2023tree} and dynamic re-planning \citep{song2023llm} to adapt to feedback.
At each turn, effective decision-making can require both short-term (context window) and long-term memory (external storage) \citep{song2023llm,huang2023memory,wang2023describe}.
Additionally, agents might need to track evolving hidden environment states influenced by prior actions \citep{ebrahimi2024your,vodrahalli2024michelangelo}.
These, and potentially more, capabilities collectively underpin robust agentic behavior.

\myparagraph{Characterizing Bottlenecks for Agents.}
Approaches to assessing model effectiveness in agentic tasks include holistic benchmarks \citep{trivedi2024appworld,patil2024gorilla} and capability-level analysis.
Benchmarks show a clear trend: larger models outperform smaller ones, prompting the question—what drives these gaps?
Few works exist that isolate bottlenecks in multi-turn tasks: reinforcement learning (RL) influence in strategic games like mazes \citep{abdulhai2023lmrl}, error taxonomies for multi-agent systems in AppWorld \citep{cemri2025multi}, and concurrent to our work, execution limits in long-horizon tasks \citep{sinha2025illusion}.
In contrast, we study the bottlenecks in procedurally-generated game environments, which allows us to inject oracle interventions.

\begin{figure*}[t]
    \centering%
    \includegraphics[width=0.99\textwidth]{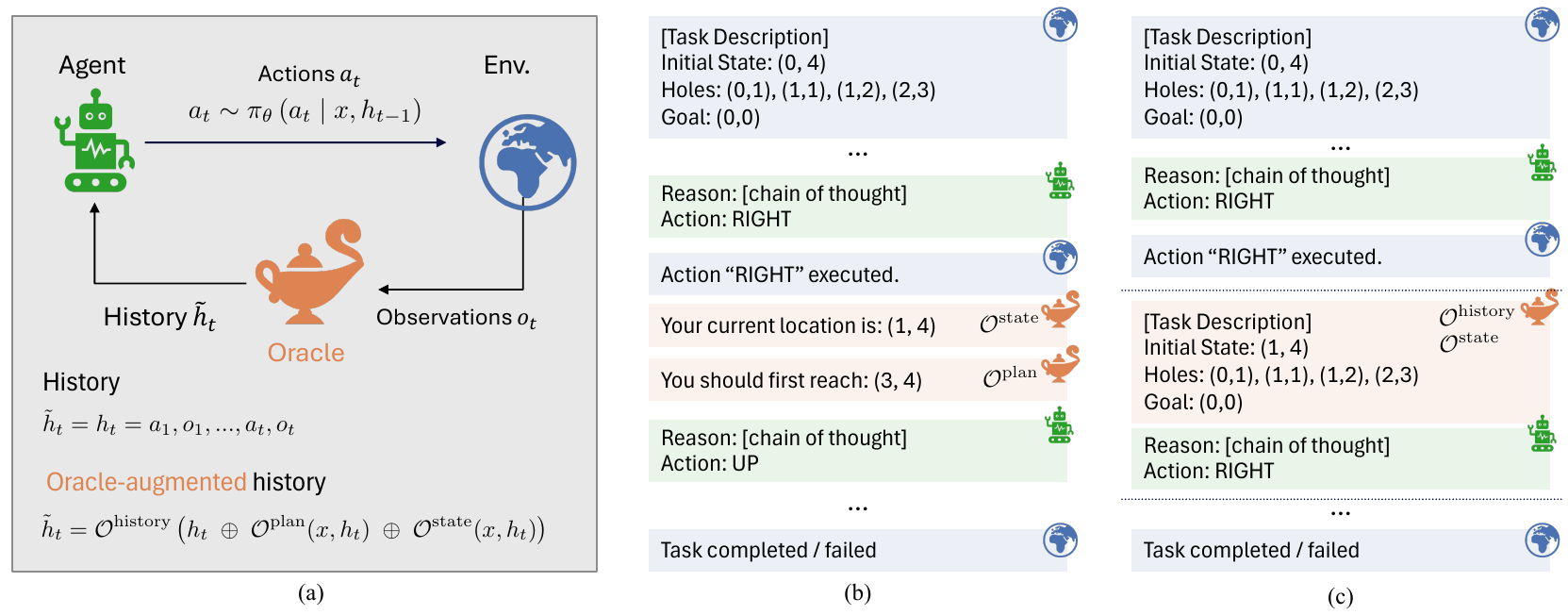}%
    \caption{\textbf{Formulation.} 
    \textbf{(a) Oracle-augmented history.} Within multi-turn tasks, we study LLM-based agents $\pi_\theta$ when additionally assisted by an oracle. We can leverage one or more oracles to modify the history $h_t$ (context for language model). 
    \textbf{(b) GridWorld example.} In this example, the agent needs to navigate from an initial 2D location to a goal location. We can use oracle $\gO^{\text{state}}$ to summarize the current location (instead of the model reflectively reasoning at each turn). Similarly, we can also use $\gO^{\text{plan}}$ to hint way points to reach the goal.
    \textbf{(c) History pruning}. Since we consider Markov decision processes, $\gO^{\text{history}}$ can be used to rewrite the task description such that the actions can be taken independent to previous steps.}
    \label{fig:oracle_formulation}
\end{figure*}

\section{Formulation: LUMINA}

In this section, we begin by detailing the underlying process that requires an agent to perform sequential decision-making to complete the task.
To better enable the agent complete the task, we then elaborate on how to augment information at each turn using oracle interventions.

\myparagraph{POMDP Tasks.}
We study tasks that can be modeled as a Partially-observable Markov Decision Process (POMDP):
$\gM = \langle \gS, \gA, \gO, T, \Omega, H , S_\mathrm{Goal}\rangle$
where $\gS$ is the hidden state space, $\gA$ the agent's action space, and $\gO$ is the observation space.
Furthermore, $T: \gS \times \gA \rightarrow \gS$ is the transition function (deterministic in our case) and $\Omega: \gS \times \gA \rightarrow \gO$ is the observation function. 
The termination can be performed either by the agent (e.g., \texttt{DONE} action) or by the environment (e.g., $t \ge \text{horizon } H$).
For simplicity, we consider a terminal reward function: the agent receives a positive reward of  $+1$ if it terminates at the goal state $S_\mathrm{Goal}$ within at most $H$ steps and receives $0$ elsewhere. The objective of the agent is to maximize the probability of success, which we refer to as the \textit{success rate} and denote by $J(\pi_\theta)$ for an agent $\pi_\theta$.

\myparagraph{Base (ReAct) Policy Agent.}
Towards taking sequential decisions to solve the task (represented as text $x$, we consider a stochastic policy modeled by a ReAct \citep{yao2023react} LLM agent $\pi_\theta$ where at each step $t$:
\begin{align*}
a_t &\sim \pi_\theta\left(\cdot \mid x, h_{t-1}\right).
\end{align*}
 Here, $h_{t-1} := (a_1, o_1, ..., a_{t-1}, o_{t-1})$ is the history of the past interactions between the agent and the environment. Action $a_t$ consists of both a chain-of-thought text (which is irrelevant to the environment) and one of the allowed operations.

\myparagraph{Oracle Interventions.}
To help our understanding and isolate factors to determine the relative role of each skill, we consider oracle interventions to assist policy $\pi_\theta$ by augmenting auxiliary information.
First, we establish the existence of an oracle $\mathcal{O}$ that, given the prompt $x$ and the context $h_{t-1}$ is able to accurately recover the belief state of the POMDP. 
Then, we consider the policy $\pi$ at every step is conditioned on \textbf{oracle-augmented history} $\tilde{h}_{t}$:
\begin{align*}
    a_t &\sim \pi_\theta\left(\cdot \mid x, \tilde{h}_{t-1}\right), \\
    \begin{aligned}
        \tilde{h}_{t-1} \\[6pt]
        {}
    \end{aligned}&
    \begin{aligned}
        = \gO^{\text{history}} &\Bigl( h_{t-1} \; \oplus \; \gO^{\text{plan}}(x, h_{t-1}) \\
                               & \; \; \oplus  \; \gO^{\text{state}}(x, h_{t-1}) \Bigr).
    \end{aligned}
\end{align*}
Generally, our oracle formulation accommodates appending ($\oplus$) a hint for the next step of an optimal plan %
(via $\gO^{\text{plan}}$) and a summary of the belief state
(via $\gO^{\text{state}}$) to the history, as well as representing the context compactly by pruning the history of redundant information (via $\gO^{\text{history}}$).
We elaborate on the details of each of these in the following paragraphs.

\myparagraph{Planning $\gO^{\text{plan}}$.} As apparent from the POMDP formulation, solving multi-turn environments requires reasoning about many steps into the future and devising a plan towards completing the task. At every step $t$, the planning intervention $\gO^{\text{plan}}(x, h_{t-1})$ is the description of a single-turn subtask. This subtask is designed not to require planning (no reasoning about the future steps needed). Most importantly, the action that accomplishes this subtask is one of the optimal actions in the environment at that moment.

\myparagraph{State Tracking $\gO^{\text{state}}$.}
Solving partially-observable long-horizon tasks requires the agent to accurately track its knowledge about the hidden state of the environment at every step.
This is highly challenging, since at each turn the agent needs to collect the information implicitly from its history of interactions with the environment and reason about the environment transitions.
Consequently, we introduce an oracle $\gO^{\text{state}}(x, h_{t-1})$ to present the precise state (e.g., current location in GridWorld) to the agent in a compact natural language form, where the state is determined from the agent's initial state and subsequent actions.
As a result, the oracle here can be interpreted as enabling the agent to exploit Markov property in the POMDP task.

\myparagraph{History Pruning $\gO^{\text{history}}$.}
It is well-known that the performance of LLMs degrade as the size of the context (history $h_t$ in our case) grows.
More relevant to us, existing work \citep{laban2025llms,vodrahalli2024michelangelo} highlights that the performance at the same task drops merely due to the presence of distractors.
This is a common problem in multi-turn LLM agentic tasks, where the history contains overcomplete information to guide the agent towards the goal, and since the size of the context (history $h_t$) grows at each turn, makes decision making more error prone.
To mitigate the influence of distractors, we consider an oracle $\gO^{\text{history}}$ that reduces the contents of the context into a compact form. In this work, we consider the simple implementation that can be done when state tracking is present. In this case, we drop the old history $h_{t-1}$, since when the compact summary $\gO^{\text{state}}(x, h_{t-1})$ is given, $h_{t-1}$ is not necessary.

\myparagraph{Step vs Task Metric $J_\mathrm{step}$.}
In multi-turn environments, the success rate can often be low because agents are required to take the correct action at each step consistently. 
This challenge is compounded by the unique difficulties inherent in such environments, such as planning and state tracking. 
For example, even if the environment involves solving single-turn step-by-step reasoning problems, the performance can suffer due to the dependency on the chain, irrespective of the absence of planning. 
However, there are cases where this setup can be advantageous by offering the agent retries; for instance, in scenarios where solving a fraction of tasks is enough for success. To measure this aspect, we define an objective over each step. We consider a step accurate if the action taken is optimal for that step. We refer to this metric as \textit{step accuracy}.

\begin{figure*}[t]
    \centering%
    \includegraphics[width=0.99\textwidth]{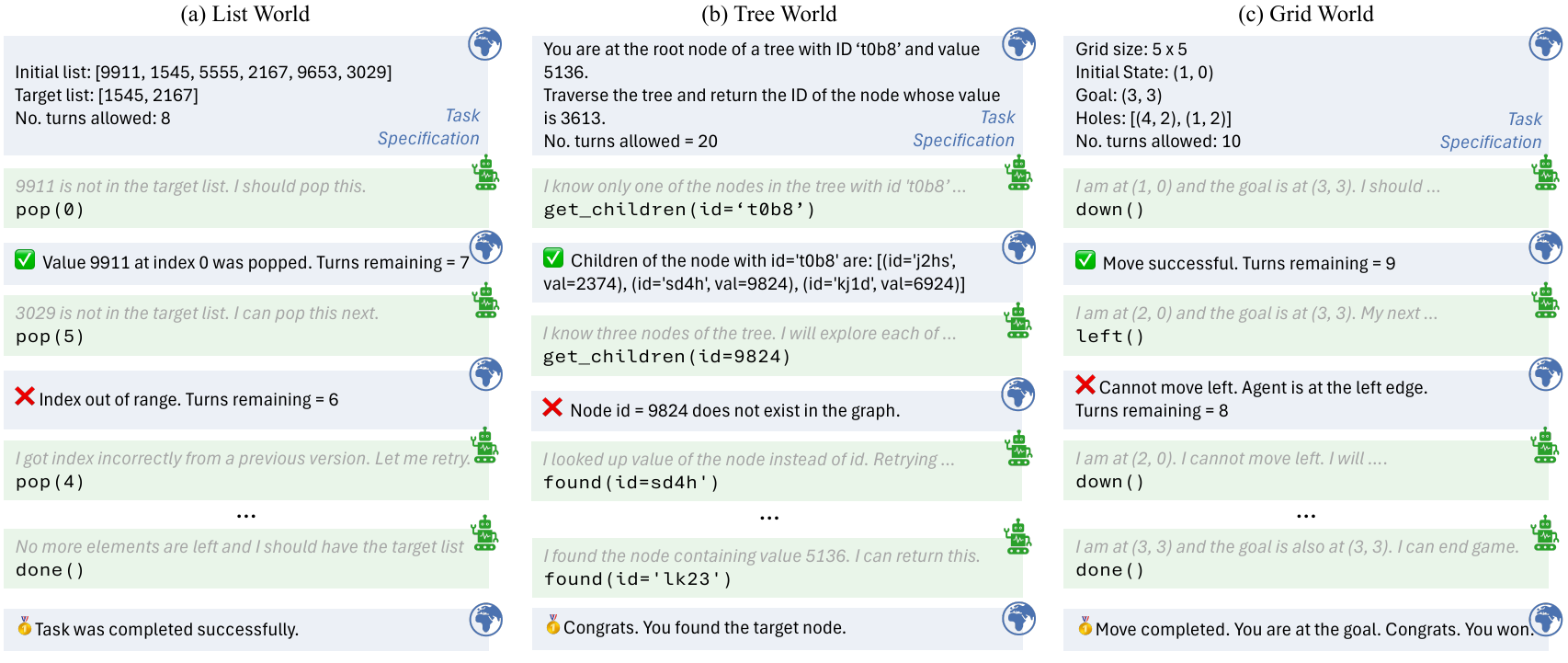}%
    \caption{\textbf{Environments.} In this work, we study the influence of oracle interventions in three unique environments. In all cases, the agent reasons (shown in gray italic) and performs an action (shown in monospace), and the environment provides minimal but sufficient feedback to help the agent progress towards the goal.
    \textbf{(a) ListWorld}: which requires modifying a python list using only \texttt{pop(idx)} actions;
    \textbf{(b) TreeWorld}: where the task is to iteratively search over a tree to find a specific node; and
    \textbf{(c) GridWorld}: where the agent needs to move from an initial location to a goal location.}
    \label{fig:environments}
\end{figure*}

\section{Experimental Results}

In this section, we first present the environments and tasks used in our experiments,
the agent and oracle setups, and conclude by reporting our empirical results and insights.

\subsection{Procedurally-generated Multi-turn Environments}

Our goal is to characterize bottlenecks of multi-turn agents by leveraging oracle interventions.
Existing benchmarks fall short for this task, since data are predominantly human-generated and are rarely accompanied with trajectory-level annotation.
Some works \citep{trivedi2024appworld} investigate marginal gains through oracle interventions, however in a very narrow scope that is admissible within the dataset.
Consequently, we propose procedurally-generated multi-turn environments with the following requirements:
(a) Minimal external knowledge: such that all necessary information can be specified in the prompt;  
(b) Simple action space: to prevent failures from constructing complex function calls;
(c) Variable complexity: to enable us analyze success by varying the complexity of the task in a procedural manner;
(d) Compositionality: such that tasks can be programmatically and accurately broken down into clear subproblems;
(e) No data contamination: since the tasks are novel and can additionally be randomly re-generated, there is little risk from contamination;
and most importantly, (f) Oracle interventions: since we know the underlying process at any instant, we can faithfully construct oracle interventions.
Figure \ref{fig:environments} shows partial examples of our environments.

\myparagraph{General Framework.}
All our environments can be cast as a POMDP. Given an initial task $x$ that can be communicated verbally, the agent needs to complete the task with a finite turn budget $T_{\max} \ge mT^* + n$, where $T^*$ is the number of actions required by an optimal policy.
To achieve the goal, the agent interacts with the environments using a simple set of actions.
All environments have a common terminating action \texttt{done}, which the agent needs to invoke once it completes the objective.
The environment provides minimal essential feedback (e.g., `move successful') at each turn.

\myparagraph{ListWorld.}
Inspired by \cite{vodrahalli2024michelangelo}, we introduce ListWorld to evaluate the ability of an LLM agent to sequentially modify and track the state of an initial object.
Specifically, the task of the agent is to prune an initial \texttt{input} Python list to a smaller \texttt{target} list.
The agent needs to prune using a single action: \texttt{pop(index)}. The agent has to pop the elements from left to right: once an element is popped, it becomes illegal to pop the elements before it. 
We control the task complexity by varying the number of elements that need to be pruned (i.e., \texttt{len(initial) - len(target)}).
To complete the task successfully, the agent at each turn needs to: 
(a) determine current list by accounting initial list and historical actions; 
(b) find the candidates to prune; and 
(c) pop the corresponding candidate.
This introduces a subtle challenge of understanding partial changes to the index-value mappings after every successful pop operation.

\myparagraph{TreeWorld.}
In this environment, we study an agent's ability to sequentially explore and gather information at each turn.
Specifically, the task is tree traversal: the agent needs to navigate from a source node to a target node to find the node containing a particular value.
The nodes (except source and target) and edges are unknown to the agent.
For simplicity and ease of analysis, we consider traversing from the root to a leaf node of a tree.
The agent needs to traverse the tree using a single action: \texttt{get\_children(node\_id)}.
Efficiently completing this task requires the agent to keep track of the frontier of unexplored nodes and sequentially explore them.
We vary the complexity of the task by controlling the number of nodes in an $m$-ary tree.

\myparagraph{GridWorld.}
We also consider a 2D grid world environment to study an agent's ability to plan and spatially navigate towards a goal.
Our grid world takes the form of an $N \times N$ grid with holes, where stepping into the holes incurs an additional cost.
The environment is fully observable, with each task stating the agent's start position and the goal position.
The agent needs to navigate to the goal using one of four actions (\texttt{up()}, \texttt{down()}, \texttt{left()}, or \texttt{right()}) and reach the goal within a specified cost budget.

\subsection{Setup: LLM Agents}
\label{sec:expt_setup}

In this section, we walk through the models we used for evaluation, how the prompts were designed to elicit closed-loop interactions with the environment, and also discuss evaluation metrics.

\myparagraph{Models.}
We focus our experiments on the Qwen3 \citep{yang2025qwen3} family of models as the policy model $\pi_\theta$.
We additionally run some experiments using Gemma 3 and GPT-4o to validate findings.
The family of Qwen models is appealing for our understanding multi-turn scenarios for two reasons:
(i) it enables us to study performance over a range of different model sizes (we focus on 4B - 32B); and
(ii) the models are already pre-trained on multi-turn interaction cycles, during the RL stage \citep{yang2025qwen3}, and hence they are well-suited for our analysis.
We report all findings by running inference in non-thinking mode (but with reasoning traces using chain-of-thought prompting) and recommended sampling temperature of 0.7.
We also ran preliminary experiments with thinking mode, but we found that token limits reduce performance.
We use the ReAct \citep{yao2023react} framework to perform the roll-outs by interleaving reasoning traces with actions.
The number of turns for which the roll-outs are performed is example dependent.
Many of our experiments are long-horizon and exceed the context length (involving contexts $>$32K).
In such scenarios, we report results using YaRN \citep{peng2023yarn} encoding, following the official recommendation.

\myparagraph{Prompting.}
Across all environments, we engineer environment-specific prompts to ensure best success of the pre-trained models.
This helps us ensure that at evaluation time, errors can be attributed with high confidence to limitations of the model rather than prompt construction.
Specifically, we:
(a) use in-context example trajectories, which demonstrate task-specific reasoning and actions;
(b) ensure the in-context examples reflect the information augmented by the oracles during roll-outs.
We found the latter to be especially important, as models underperformed if the format of in-context examples did not appear consistent with environmental feedback during roll-out.

\myparagraph{Evaluation Metrics.}
Our primary evaluation metric is the \textit{success rate}, i.e., whether the model completes the task within the specified horizon budget.
In addition, we also report the \textit{step accuracy} i.e., whether the agent's action at a particular step in a multi-turn trajectory is optimal.
An action is considered optimal, if it is aligned with at least one optimal policy (of multiple non-unique policies) of the current state.

\subsection{Setup: Oracle Interventions}
To better understand the impact of state tracking, planning, and growing prompt length of multi-turn environments, we conduct a thorough evaluation of the models across all sizes in the presence or absence of our oracle interventions. Each intervention removes one of the aforementioned challenges, and their impact on the agent's performance allows us to understand the importance of each one. 

In all the environments, the oracle interventions are designed either to \textit{augment} the information provided  or \textit{truncate} to sufficient information necessary to achieve the goal.
Planning and state tracking interventions augment the context and can be activated or deactivated independently.
In contrast, history pruning depends on the state tracking intervention, as it removes all preceding steps from the context, making it impossible for the agent to track the state.
In the cases where state tracking intervention is active, the provided state contains sufficient information for optimal decision-making.
Therefore, we have six possible configurations for the oracle intervention.

\begin{figure*}[t]
  \centering
    \includegraphics[width=\textwidth]{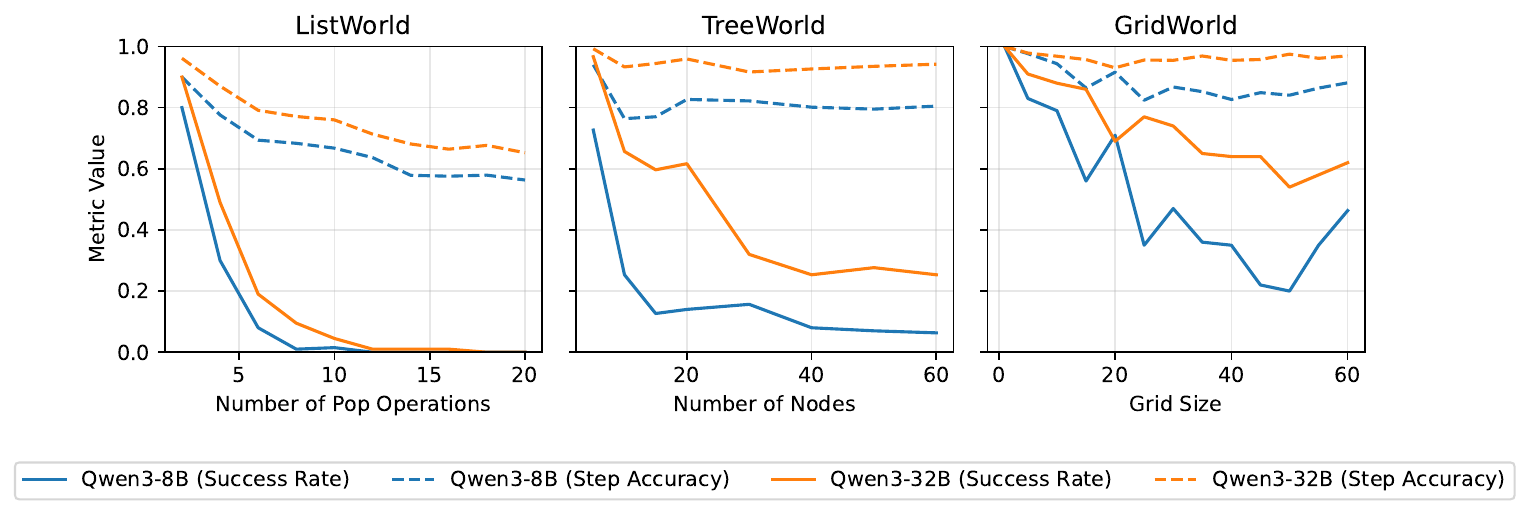}
    \caption{Success rate and step accuracy of Qwen3-8B and Qwen3-32B models by task horizon in ListWorld \textit{(left)}, TreeWorld \textit{(middle)}, and GridWorld \textit{(right)}.}
    \label{fig:horizon_vs_success}
\end{figure*}

\begin{figure*}[t]
  \centering
  \begin{subfigure}[b]{0.85\textwidth}
    \includegraphics[width=\textwidth]{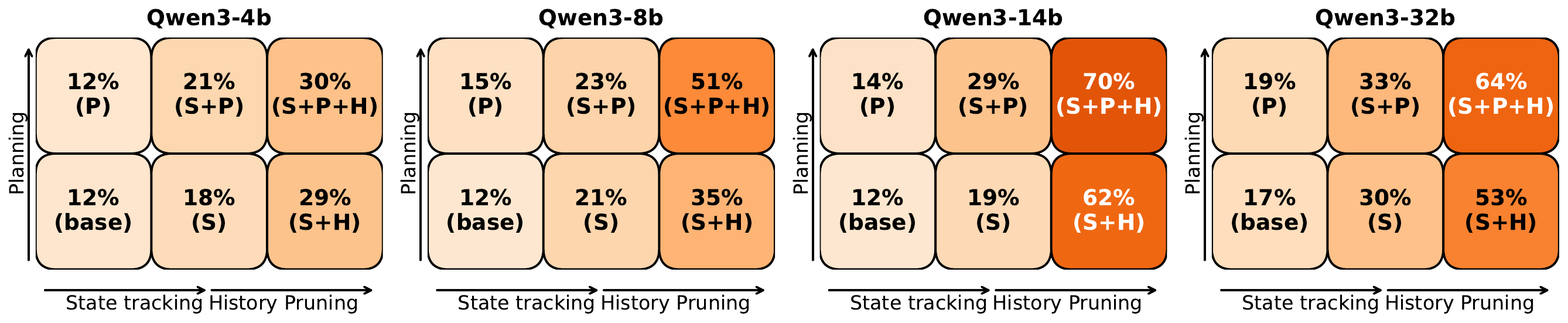}
    \label{fig:grid_listworld}
  \end{subfigure}
  \begin{subfigure}[b]{0.85\textwidth}
    \vspace{-0.35cm} %
    \includegraphics[width=\textwidth]{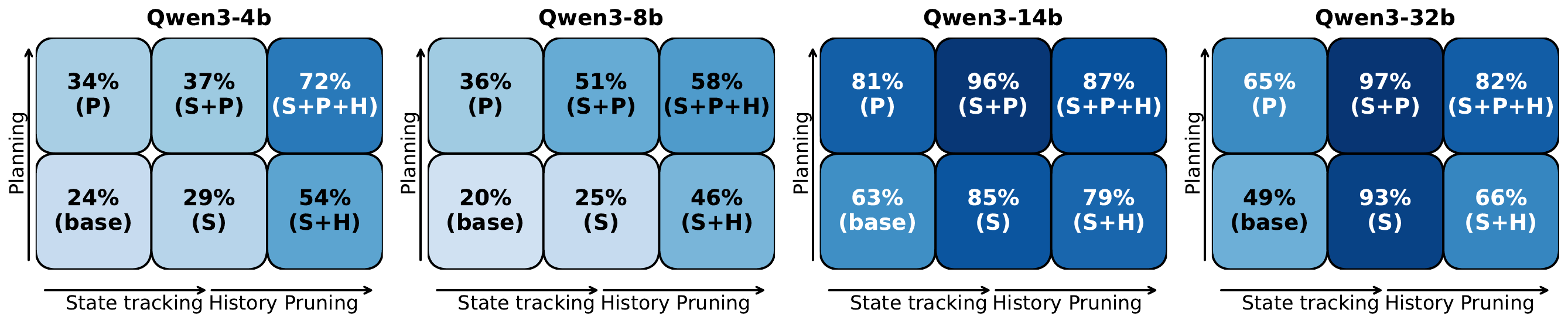}
    \label{fig:grid_treeworld}
  \end{subfigure}
  \begin{subfigure}[b]{0.85\textwidth}
    \vspace{-0.35cm} %
    \includegraphics[width=\textwidth]{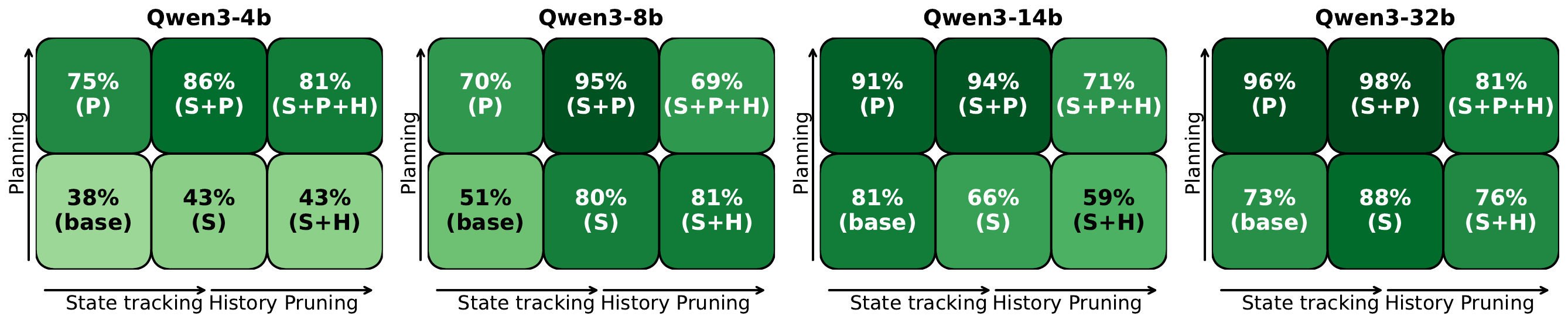}
    \label{fig:grid_gridworld}
    \vspace{-0.35cm}
  \end{subfigure}
      \caption{Influence of oracle interventions on ListWorld \textit{top}, TreeWorld \textit{(middle)}, and GridWorld \textit{(bottom)}. Results are averaged over all horizon lengths. The labels indicate the active oracles (S: state tracking, P: planning, and H: history pruning).}
      \label{fig:intervention_influence}
      \vspace{-1em}
\end{figure*}

\subsection{Results and Analysis}
\label{sec:expt_results}

We examine the performance of Qwen3 models on our environments. Fig.~\ref{fig:horizon_vs_success} shows the success rate of the Qwen3 8B and 32B models on our three environments as a function of the task horizon. We observe that despite the strong performance on problems with a short horizon, the success rate drops drastically as we increase the horizon. This is aligned with the notoriously challenging nature of long-horizon tasks. We study the driving forces behind this phenomenon with our framework. 

\myparagraph{Overall Oracle Impact.} 
Fig.~\ref{fig:intervention_influence} provides the success rate of 4B, 8B, 14B, and 32B models in each environment for all six oracle configurations, averaged over complexity levels.
The addition of each intervention improves the success rate in most cases.
The exact impact of each intervention varies among the environments and model sizes.

\begin{figure*}[t]
    \centering
  \begin{subfigure}[b]{0.85\textwidth}
    \vspace{-0.35cm}
      \includegraphics[width=\textwidth]{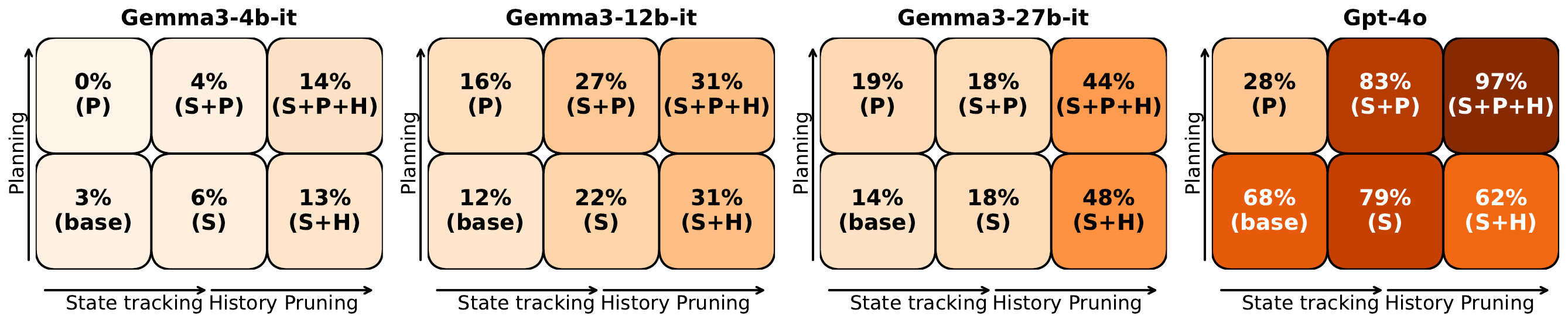}
      \label{fig:grid_listworld_extra}
    \vspace{-0.35cm}
  \end{subfigure}
      \caption{Influence of oracle interventions for Gemma 3 and GPT-4o on ListWorld. Results are averaged over all horizon lengths. The labels indicate the active oracles (S: state tracking, P: planning, and H: history pruning).}
      \label{fig:gemma_gpt4o}
      \vspace{-1em}
\end{figure*}

\begin{figure}[th]
    \centering
    \begin{subfigure}[b]{0.48\textwidth}
        \includegraphics[width=\textwidth]{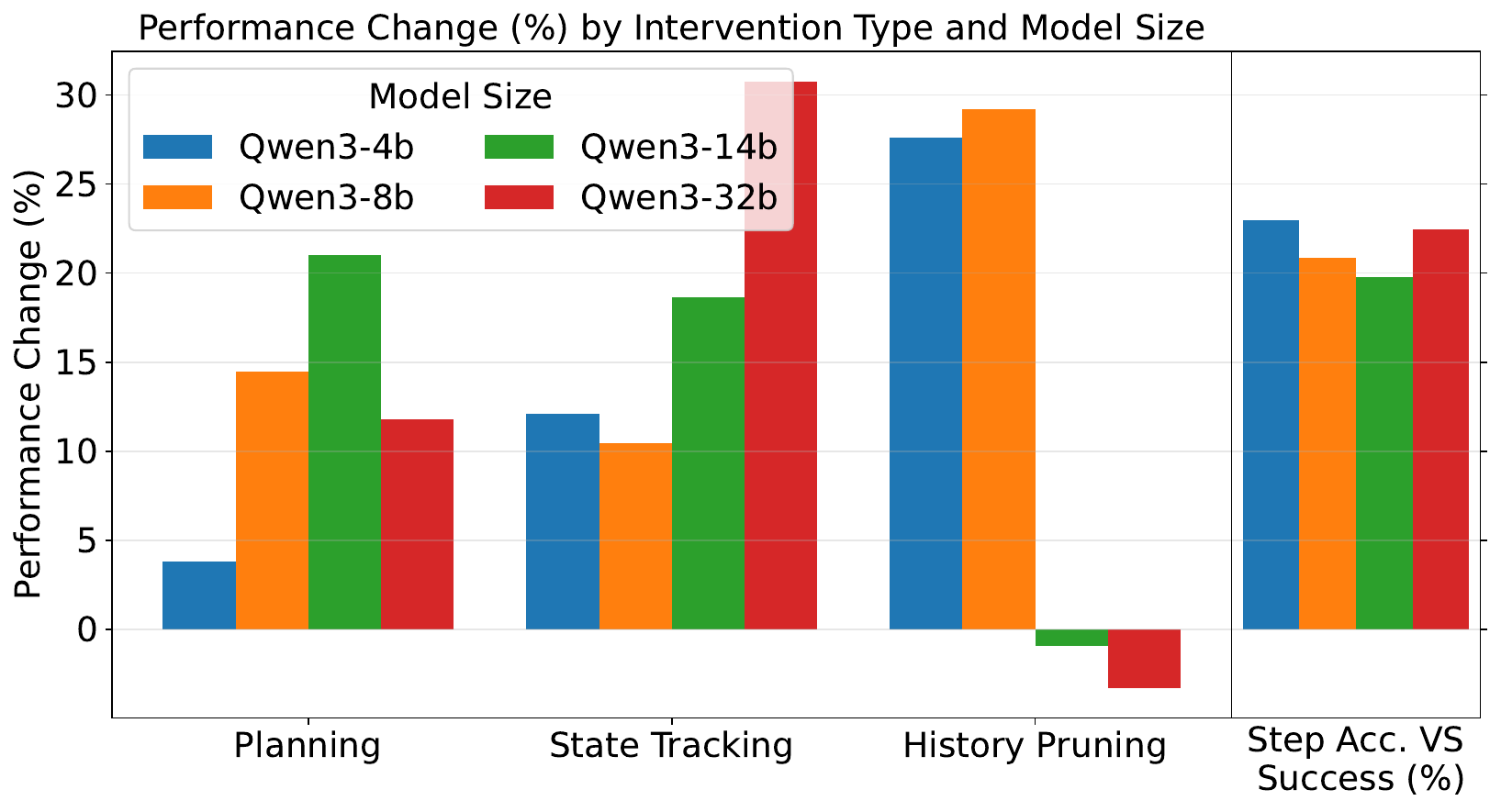}
        \caption{Model Size}
        \label{fig:ivsse_modelsize}
    \end{subfigure}
    \hfill
    \begin{subfigure}[b]{0.48\textwidth}
        \includegraphics[width=\textwidth]{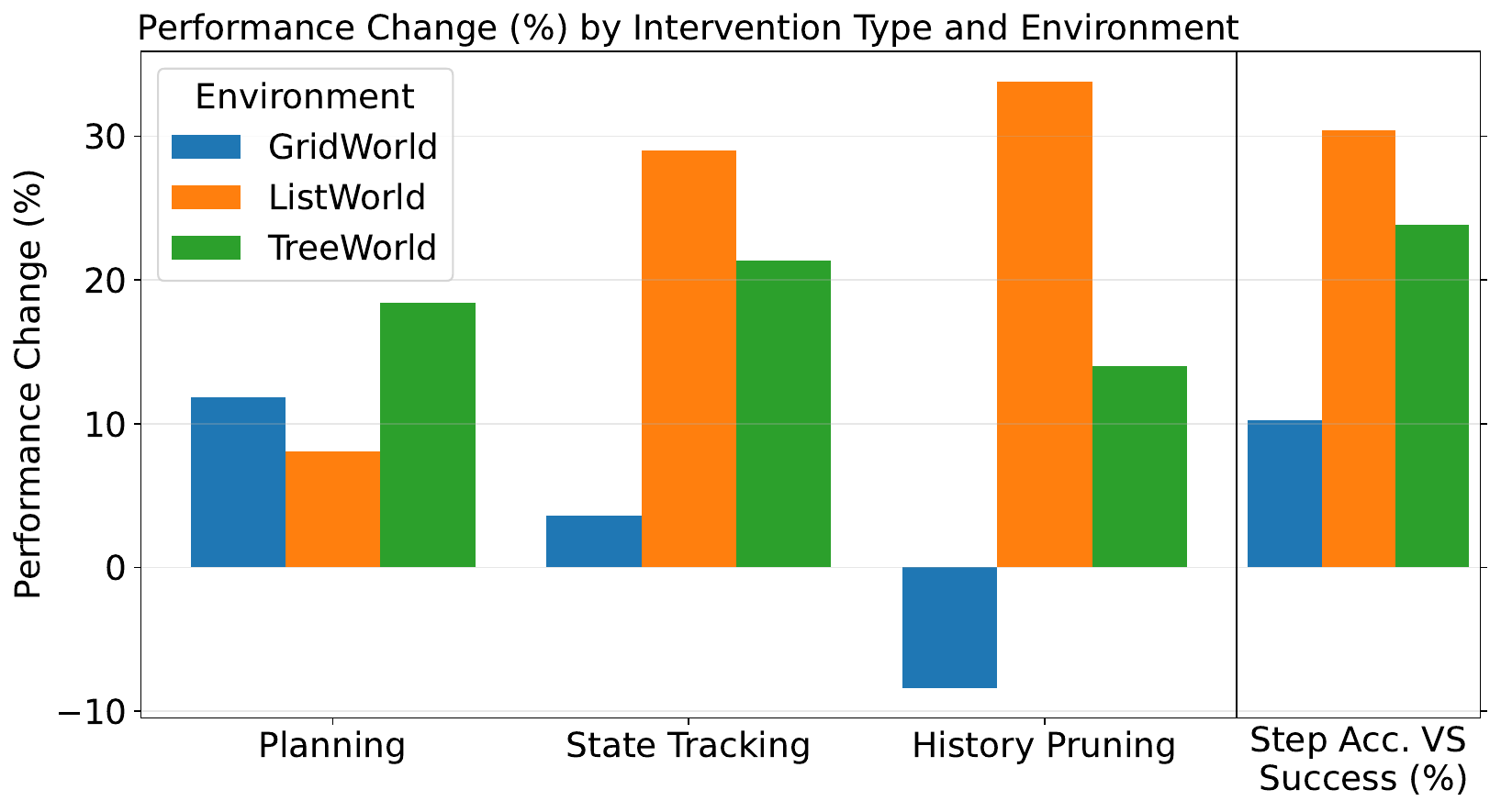}
        \caption{Environment}
        \label{fig:ivsse_env}
    \end{subfigure}
    \caption{\textbf{Performance change from interventions.} The impact of each intervention and varies depending on the model size (\subref{fig:ivsse_modelsize}) and environment (\subref{fig:ivsse_env}). The relationship between step accuracy and success (\%) also varies.}
    \label{fig:intervention_vs_size_and_env}
    \vspace{-1em}
\end{figure}

\myparagraph{Model Scale Shifts What Matters.}
Larger models outperform smaller models in every environment, and consequently, exhibit less potential for improvement through interventions.
To enable an insightful comparison across model sizes, we adjust for this natural imbalance by
picking a longer horizon for larger models and effectively leveling the success rates.
For ListWorld and TreeWorld, we choose settings where models succeed about $30\%$ of the time.
In GridWorld, larger models never reach this low success rate, and we pick the complexity such that success rate becomes about $75\%$ for all model sizes.
Figure~\ref{fig:ivsse_modelsize} shows the impact of the interventions on the success rate for each model size after the leveling selection.
The results are averaged over the choice of environment and other interventions. 
The impact of planning intervention and the difference between task success rate and step accuracy seem generally unaffected by the model size.
History pruning is effective for the 4B and 8B models, while the 14B and 32B even suffer from this removal.
Another observation is that larger models tend to benefit more from state tracking.
This indicates that improving these capabilities in models will have largerly different impact at different model scales.

\myparagraph{Rules Dictate the Challenges.}
Figure~\ref{fig:ivsse_env} compares the impact of each intervention in different environments.
Again, we choose the task complexity to level the success rates and average over the choice of model size and other interventions.
The oracles' impact significantly varies among environments.
ListWorld performance is boosted best by history pruning, indicating that it is challenging for the agent to attend to the relevant information.
Given GridWorld's navigational nature, knowing where to go and also knowing where one comes from is intuitively useful information.
This may explain why planning improves while history pruning deteriorates GridWorld performance.
While the state tracking oracle has little impact in GridWorld, it is the most impactful oracle in TreeWorld, likely due to complex back tracking required for the task.

\myparagraph{Every Step Counts.}
Our experiments reinforce the hypothesis that the main challenge of multi-turn environments is that success requires \textit{consistently} taking correct actions over many steps.
The dashed lines in Fig.~\ref{fig:horizon_vs_success} show the step accuracy of the model for each value of task horizon.
The step accuracy is larger than the task success rate, and even in cases where the success rate is almost zero, a step accuracy above $60\%$ indicates that the majority of the steps are correct.
However, as the number of required steps increases, the agent becomes more likely to fail on the task due to the occasional wrong actions it takes.
In order to solve these tasks reliably, the agent needs to be almost perfect at all steps.
In fact, environments like ListWorld are irrecoverable and a single step leads to failure.
The large discrepancy between step accuracy and success rate in Fig.~\ref{fig:ivsse_env} is likely a result of that environment property.

\myparagraph{Results with Different Models.}
In Fig.~\ref{fig:gemma_gpt4o}, we provide the success rate of GPT-4o \citep{hurst2024gpt} and Gemma 3 models \citep{team2025gemma} on ListWorld for different oracle configurations.
Gemma 3 results are aligned with the findings from the Qwen3 models.
Planning intervention shows minimal benefits while state tracking and history pruning significantly help the model.
Both Qwen3 and Gemma 3 models across all sizes benefit from history pruning in ListWorld, but GPT-4o, which is much larger, starts to show signs of performance degradation with history pruning.
This further confirms our finding that history pruning benefits smaller models but can hurt larger models.

\section{Conclusion}
In this paper, we worked towards understanding the discrepancy LLMs excelling at complex single-turn reasoning tasks while underperforming in multi-turn closed-loop feedback-driven tasks.
Key to our approach was to analyze the discrepancy in terms of skills that are required in agent-specific use cases, such as planning, tracking state, and history pruning.
To this end, we proposed a counterfactual framework with oracle interventions, where the oracle augment and prune the information exposed to the agent at each turn.
To enable oracle interventions, we additionally propose three procedurally-generated environments, which lets us control task complexity, and more importantly, we can at any turn estimate the set of optimal actions that can be used to guide the agent.
Our findings indicate that while oracle skills enable the LLM policy to generally improve, the effectiveness of each skill is also significantly influenced by the model size and the environment.
Further, our experiments highlight the importance of close to perfect step accuracy to avoid compounding errors.
These insights may guide future development to robust LLM agents for multi-turn tasks.

\subsubsection*{Limitations}
This paper presents the first step towards explaining the performance degradation of capable LLMs in multi-turn long-horizon agent tasks.
While we find valuable insights, many important steps remain to fully understand the performance degradation.
First, we rely on prompt-based mechanisms to elicit agent-like behavior.
While such mechanisms have been shown to be a strong baseline, performance are also influenced by the prompt itself and it is unlikely to find one prompt construction that is optimal for all models, sizes, and environments.
This may potentially be alleviated by post-training given specific prompts.
Second, we introduce oracle information by augmenting or rewriting the instructions in context visible to the LLM, and hence rely on the LLM to accurately follow the instruction.
Although LLMs generally show remarkable instruction following capabilities, it is nonetheless imperfect and a framework to additionally enable an `instruction following' oracle would strengthen the analysis.
Third, we run our analysis on simple programmable environments and benefit from being able to accurately and efficiently construct and isolate oracle interventions.
While it is insightful to study this on real-world applications, annotating oracle is often intractable or ill-defined.
We do not anticipate immediate risks arising from our work, but we do acknowledge that autonomous agentic system may pose a big threat if faulty, unreliable, or used maliciously.
Therefore, we believe that analyses, such as ours, are crucial to mitigate such risks in the future.

\bibliography{acl_latex}

\newpage
\clearpage

\appendix

\section{Prompts}

\subsection{Prompt: ListWorld}

We used the following prompt for ListWorld experiments

\begin{lstlisting}
You are an assistant designed to **modify lists** through a sequence of simple commands that you can execute at every turn. You will be given an initial list and a target list. Your task is to modify the initial list to a target list using the following functions, provided in JSON format:
```json
{
    "pop": {
        "name": "pop",
        "type": "function",
        "description": "Removes an element from the environment's list. Index of the elements after the removed one will be reduced by 1.",
        "parameters": {
            "type": "object",
            "properties": {
                "id": {
                    "type": "integer",
                    "description": "The ID of the element to remove."
                }
            },
            "required": [
                "id"
            ]
        }
    },
    "done": {
        "name": "done",
        "type": "function",
        "description": "Terminates the task. Should be called when no more operations are needed.",
        "parameters": {
            "type": "object",
            "properties": {}
        }
    }
}
```
The initial list has all the elements of the target list in the same order, but contains some extra elements that need to be removed. To remove them, you should use the 'pop' function with the index of the element you want to remove. Remember that once an element is removed, the index of elements after it will decrease by 1. The most important rule is that you can **only pop the elements from left to right**. Once you pop an element, the elements before it (thoses with a smaller index) can no longer be removed, and you will get an error if you try to do so. Once you are done and have turned the initial list to the target list, you should call the 'done' function.

You must reason step-by-step, choosing actions based on the current state of the list. Avoid redundant queries and aim for efficiency.
Provide your response as a single python function call enclosed in a code block:
```python
function_call(arg1=val1, arg2=val2, ...)
```

=== Starting new task ===
Initial list: ${initial_list}
Target list: ${target_list}
Your task is to modify the initial list to target list by calling the 'pop' function. Call 'done' function once you are done. Remember, **only pop the elements from left to right**.
\end{lstlisting}

\subsection{Prompt: TreeWorld}

We used the following prompt for tree world:
\begin{lstlisting}
SYSTEM_PROMPT = f""""You are a reasoning agent searching a tree for a node with a specific value (which may or may not be reachable by you). Each node has two attributes: (1) "id" (unique string) and (2) "value" (unique integer). In each task, you are provided with a partial information about some of the nodes in the tree. For each node included in this information, you are given the id and the value. For some of the nodes the list of the node's children ids is also provided. The format in that case is: (id=<node_id>, value=<node_value>) -> [<child_id1>, <child_id2>, ...]. Note that an empty list indicates that the node is a leaf and has no children. For other nodes, the children are not given to you. In that case, you are given: \"UNKNOWN\" in place of the children ids.

A target value will be given to you at the beginning of each task. Your job is to try to find a node with this value and report its id. You should do this using the following functions, provided in JSON format:
```json
{
    "get_children": {
        "name": "get_children",
        "type": "function",
        "description": "Returns the list of children nodes for a given node ID (i.e., the outgoing edges)",
        "parameters": {
            "type": "object",
            "properties": {
                "id": {
                    "type": "string",
                    "description": "The ID of the node whose children are to be retrieved."
                }
            },
            "required": [
                "id"
            ]
        },
        "returns": {
            "type": "array",
            "items": {
                "type": "object",
                "properties": {
                    "id": {
                        "type": "string"
                    },
                    "val": {
                        "type": "integer"
                    }
                },
                "required": [
                    "id",
                    "val"
                ]
            },
            "description": "List of child nodes as objects with 'id' and 'val'."
        }
    },
    "found": {
        "name": "found",
        "type": "function",
        "description": "Indicates that the node with the specified ID contains the target value.",
        "parameters": {
            "type": "object",
            "properties": {
                "id": {
                    "type": "string",
                    "description": "The ID of the node that contains the target value."
                }
            },
            "required": [
                "id"
            ]
        },
        "returns": {
            "type": "string",
            "description": "Confirmation that the node with the given ID contains the target value."
        }
    },
    "unreachable": {
        "name": "unreachable",
        "type": "function",
        "description": "Indicates that the node with the target value is impossible to reach.",
        "parameters": {
            "type": "object",
            "properties": {}
        },
        "returns": {
            "type": "string",
            "description": "Confirmation that the target value was not possible to find in the tree."
        }
    }
}
```
You can ask for the ids and values of the children of a node by calling the `get_children` function with the node's id.  
If you find the target node (the node with the target value), return its id using the `found` function. After calling `found` the task will terminate and you succeed if you have reported the correct id. 
If you believe it is impossible to find the target node, call the `unreachable` function.  After calling `unreachable` the task will terminate and you succeed if it was impossible for you to find the target node.

You must reason step-by-step, choosing actions based on the current state of the search. Avoid redundant queries and aim for efficiency.
Provide your response as a single python function call enclosed in a code block:
```python
function_call(arg1=val1, arg2=val2, ...)
```

=== Starting new task ===
Your task is as follows: Find the id of the node with value **${target_node_val}**
Once you find the target node containing this value, return its id by calling `found` function. If you think it is impossible to find this node, call the `unreachable` function.
Provide all responses as a single python function call enclosed in a code block.

\end{lstlisting}

\subsection{Prompt: GridWorld}

We used the prompt below for GridWorld:

\begin{lstlisting}
You are an intelligent agent playing a grid world navigation game. Your goal is to move from the given start position to the goal position using the fewest possible moves. The game board is a 2D grid with the following properties:

- The top-left corner is coordinate (0, 0), and the bottom-right corner is (size-1, size-1).
- You will be given:
    * The size of the board (N x N)
    * Your starting position (row_index, column_index)
    * The goal position (row_index, column_index)
    * A list of hole positions (each a coordinate)
    * The maximum number of moves allowed
- You can move using these actions: `up()`, `down()`, `left()`, `right()`
- *Only* if you have reached the goal, call `done()` to terminate the game. Once you terminate the game, you are not allowed any more moves.
- You can reason, but always end by specifying a single action within triple fenced blocks. Example 
```python
up()
``` 
or 
```python
done()
``` 
- Each move costs **1 move**.
- If you move into a hole, you incur a **penalty of 3 additional moves** (because it is hard to get out of a hole).
- You must stay within the grid boundaries.
- Your objective: **Reach the goal in as few moves as possible without exceeding the maximum allowed moves.**
- After each move, you will receive the updated position and remaining moves.
- In the triple fenced blocks, do not write anything except the next action in the required format.

=== Your Task ===
The grid world game is set up as follows:
- Board size: ${size} x ${size}
- Start position: ${start}
- Goal position: ${goal}
- Holes at: ${holes}
- Your move budget is: ${max_moves}

Your task: Navigate from the start to the goal using the fewest moves possible. Remember:
- You can move using the following actions: `up()`, `down()`, `left()`, `right()`
- If you reached the goal, terminate by performing action `done()`
- Each action must be in a triple-fenced Python code block, like: 
```python
right()
```
- Avoid holes if possible, as they cost extra moves.
- Do not exceed the maximum allowed moves.

Begin your first move now.
"""
\end{lstlisting}

\end{document}